\documentclass{pnastwo}

\usepackage{graphicx}

\usepackage{amssymb,amsfonts,amsmath}

\contributor{A Brief Report}
\url{}
\copyrightyear{2014}
\issuedate{Issue Date}
\volume{Volume}
\issuenumber{Issue Number}

\begin{document}


\title{Model Predictive Control for Micro Aerial Vehicle (MAV) Systems } 






\author{Gautham Vasan\affil{1}{NIT Trichy},
Arun Kumar Singh\affil{2}{IIIT Hyderabad}
\and
K. Madhava Krishna\affil{1}{IIIT Hyderabad}}

\contributor{A Report of the Project at IIIT Hyderabad}


\maketitle 

\begin{article}


\begin{abstract}
This paper presents a method for path-following for quadcopter trajectories in real time. Non-Linear Guidance Logic is used to find the intercepts of the subsequent destination. Trajectory tracking is implemented by formulating the trajectory of the quadcopter using its jerk, in discrete time, and then solving a convex optimization problem on each decoupled axis. Based on the maximum possible thrust and angular rates of the quadcopter, feasibility constraints for the quadcopter have been derived. In this report we describe the design and implementation of explicit MPC controllers where the controllers were executed on a computer using sparse solvers to control the vehicle in hovering flight. 
\end{abstract}





\abbreviations{MPC, Model Predictive Control; NLGL, Non-Linear Guidance Logic; VTP, Virtual Target Point}



\section{I. Introduction}

\dropcap{D}ue to their remarkable agility and robustness, Quadcopters have become an exciting area of research and a common sight among aircraft hobbyists, academic researchers and industries. In general, they have high thrust to weight ratio which allows them to carry significant payloads or have high lateral accelerations when not carrying a load.  Quadcopters are nowadays being used to study a plethora of concepts ranging from non-linear control[1] and learning[2] to vision-based pose estimation[3].\newline
\newline
When Quadcopters are used for tasks like mapping, search and rescue, patrol, surveillance and product delivery, they're required to follow a predefined path at a particular height. In general, circular and straight line paths are used. Path-following algorithms are used to ensure that the Quadcopter will follow a predefined path in two or three dimensions at constant height. Another important consideration is that the path-following algorithms are robust and accurate with regard to turbulences. A wide range of path-following algorithms have been proposed [4]. 
\newline

In this paper we use the Non-Linear Guidance Logic (NLGL) [4] coupled with Model-Predictive Control [5] for trajectory tracking for the Quadcopter. The main objective of this approach includes:

\begin{itemize}
  \item The trajectories must be feasible under the dynamic and input constraints of the Quadrotor
  \item The generated trajectories should bring the vehicle to the target position on the path as quickly as possible,
  \item The trajectory calculation must be fast enough to be used online, and 
  \item It must be possible to generate an implicit feedback control law by re-planning the trajectory intercept at each controller update, and applying the control inputs of the first section of it. 
\end{itemize}

A scheme is presented in [6] for generating state interception trajectories for Quadcopters; that is, trajectories starting from an arbitrary state and achieving a (reduced) end state in a specified amount of time, whilst satisfying input constraints.\newline 
 
This approach builds on the previous work presented above in that we seek to develop that satisfies various constraints and is also fast enough to allow real-time planning. Trajectory tracking is done by formulating the trajectory of the quadrocopter in its jerk, in discrete time, and then solving a convex optimization problem on each decoupled axis. The NLGL algorithm is now used to track the path effectively.\newline

The quadrocopter model is presented in Section II, with the trajectory tracking scheme given in Section III. Section IV discusses simulation results, and an outlook is given in Section V. 


\section{II. Dynamic Model}

The Quadcopter is modeled as a rigid body with six degrees of freedom: linear translation along the inertial x1, x2 and x3 axes, and three degrees of freedom describing the rotation of the frame attached to the body with respect to the inertial frame, which is taken here to be the proper orthogonal matrix R. The control inputs to the system are taken as the total thrust produced f , for simplicity normalized by the vehicle mass and thus having units of acceleration; and the body rates expressed in the body-fixed frame as ω = (ω1 , ω2 , ω3 ). These are illustrated in Fig. 1.

\begin{figure}
  \centering
    \includegraphics[width=0.5\textwidth]{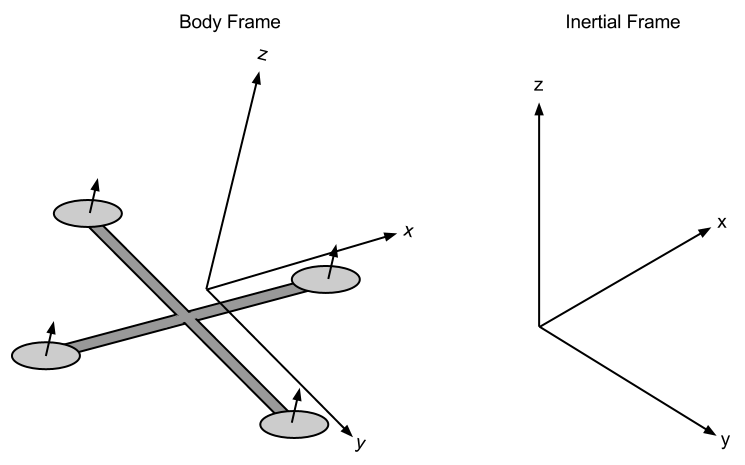}
  \caption{The Inertial Coordinate System}
  \label{fig:3d}
\end{figure}
Figure~\ref{fig:3d}

The mixing of these inputs to individual motor thrust commands is done on board the vehicle, using feedback from gyroscopes. It is assumed that the time constant of the onboard controllers is low enough to have negligible influence on the algorithm presented here. Because of their low rotational inertia, Quadcopters can achieve extremely high rotational accelerations (on the order of 200 rads−2 [7]) about the ω1 and ω2 axes, while it will be shown that the rotation about ω3 is not needed for the trajectories considered here.\newline

The differential equations governing the flight of the quadcopter are now taken as those of a rigid body [8]

\begin{equation} 
    \ddot x = Re_3f + g  
\end{equation}
 
\begin{equation} 
     \dot R = R [[wx]]
\end{equation}

with e3 = (0,0,1) and [[wx]] the skew symmetric matrix form of the vector cross product such that:
 
\begin{equation} 
      [[wx]] = \left| \begin{array}{ccc}
- w_3 & -w_2 & 0 \\
w_3 & 0 & -w_1 \\
-w_2 & w_1 & 0 \end{array} \right| 
\end{equation}

and g = (0, 0, −g) the acceleration due to gravity. Note the
distinction between the vector g and scalar g.

\subsection{A. Formulation in terms if Jerk\newline}We follow [6],[9] in considering the trajectories of the quadrocopter in terms of the jerk of the axes, allowing the system to be considered as a triple integrator in each axis and simplifying the trajectory generation task. It is assumed that a thrice differentiable trajectory x(t) is available, where the jerk is written as:

\begin{equation} 
    j = \dddot x = (\dddot x_1, \dddot x_2, \dddot x_3) 
\end{equation}
 
The input thrust f is then found by applying the Euclidean
norm · to,

\begin{equation} 
    f = || \ddot x - g||
\end{equation}

\subsection{B. Feasibility and constraints of the decoupled axis \newline}
A quadrocopter trajectory described by (1) and (2) is considered to be feasible if the thrust and the magnitude of the body rates lie in some feasible set of values, defined as

\begin{equation} 
  0 < f_{min} \leq f \leq f_{max} 
\end{equation}

Note that \(fmin > 0 \) for fixed-pitch propellers with a fixed direction of rotation, and, specifically, that the requirement on the thrust input is non-convex. 

\begin{equation} 
 f_{min}^2 \leq \ddot x_1^2 + \ddot x_2^2 + \ddot x_3^2 
\end{equation}

The following conservative box constraints are applied to yield convex constraints:

\begin{equation} 
 \ddot x_{min{1}} = -\ddot x_{max{1}} \leq \ddot x_1 \leq x_{max{1}}
\end{equation}
 
\begin{equation} 
 \ddot x_{min{2}} = -\ddot x_{max{2}} \leq \ddot x_2 \leq x_{max{2}} 
\end{equation}

\begin{equation} 
 \ddot x_{min{3}} = \ddot f_{min} -g \leq \ddot x_3 \leq x_{max{3}} 
\end{equation}

The resulting trajectories are guaranteed to be feasible with respect to the thrust limit if

\begin{equation} 
 \ddot x_{max{1}}^2 + \ddot x_{max{2}}^2 + \ddot x_{max{3}}^2 \leq f_{max}^2 
\end{equation}

\section{III. Trajectory Tracking}

The three axes are decoupled and hence the optimization problem is solved separately for each axis. The trajectory tracking is rewritten as an optimal control problem, with boundary conditions defined by the quadrocopter’s initial and (desired) final states. The cost function to minimize is chosen as 

\begin{equation} 
J_x = \sum\limits_{i=1}^5 (x_{target} - x_i)^2   
\end{equation}

\begin{equation} 
J_y = \sum\limits_{i=1}^5 (y_{target} - y_i)^2 
\end{equation}

\begin{equation} 
J_z = \sum\limits_{i=1}^5 (z_{target} - z_i)^2  
\end{equation}

A Nonlinear Guidance Law based method described in [4] is used to determine the destination intercept along the given path. For simplicity, assume the UAV is following a path, as shown in Figure 4. At the current UAV position p, draw a circle of radius L. The circle will intercept the path at two points q and ql . Depending on the direction in which the UAV has to move, either q or \( q_1 \) . will be selected. The advantage of this algorithm is that the same algorithm can be applied to any type of trajectory. The stability of the guidance law is shown using Lyapunov stability arguments[].\newline 

Once the target intercepts are calculated, future position, velocity and acceleration of the quadcopter are predicted using convex optimization and having constraints based on the quadcopter model. The discrete time formulation (described in the next sub-section) is used to predict up to 5 values if position, velocity and acceleration. The first value of the jerk calculated from the constraints is given as input.

\begin{figure}
  \centering
    \includegraphics[width=0.5\textwidth]{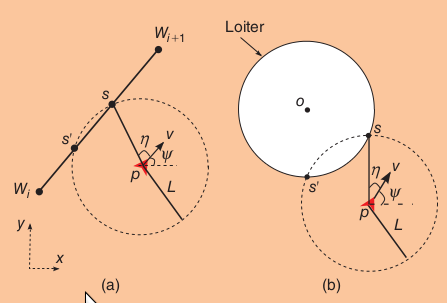}
  \caption{Determining the virtual target point (VTP) for NLGL. The VTP for (a) the straight-line path and (b) the loiter path.
}
  \label{fig:VTP}
\end{figure}
Figure~\ref{fig:VTP}

\subsection{A. Discrete Time Formulation \newline}
The trajectory generation problem for each decoupled axis is rendered finite dimensional by discretizing the time with uniform steps of size ∆t. Each axis is then a discrete time linear, time invariant system in the state z, consisting of position, velocity and acceleration, with scalar jerk input \( j = \dddot x\), where the axis subscripts have been neglected for
convenience.

\begin{equation} 
 j[k] = \dddot x(k\Delta t)
\end{equation}

\begin{equation} 
z[k] = [x(t_1) x(t_2) . . . . x(t_n) ]^T 
\end{equation}

\begin{equation} 
 \left[ \begin{array}{c} x(t_1) \\ x(t_2) \\ x(t_3) \\ x(t_4) \\ x(t_5) \end{array} \right] = \begin{bmatrix} \Delta \frac{t^3}{6} & 0 & 0 & 0 & 0\\ 
\Delta \frac{t^3}{6} & \Delta \frac{t^3}{6} & 0 & 0 & 0 \\ 
\Delta \frac{t^3}{6} & \Delta \frac{t^3}{6} & \Delta \frac{t^3}{6} & 0 & 0 \\ 
\Delta \frac{t^3}{6} & \Delta \frac{t^3}{6} & \Delta \frac{t^3}{6} & \Delta \frac{t^3}{6} & 0 \\ 
\Delta \frac{t^3}{6} & \Delta \frac{t^3}{6} & \Delta \frac{t^3}{6} & \Delta \frac{t^3}{6} & \Delta \frac{t^3}{6} \end{bmatrix}
\times \left[ \begin{array}{c} u_1 \\ u_2 \\ u_3 \\ u_4 \\ u_5 \end{array} \right]  
\end{equation}

\[
+  \left[ \begin{array}{c} \Delta \frac{t^2}{2} \\ 2\Delta \frac{t^2}{2} \\ 3\Delta \frac{t^2}{2} \\ 4\Delta \frac{t^2}{2} \\ 5\Delta \frac{t^2}{2} \end{array} \right] \times \ddot x(t_1)
+ \left[\begin{array}{c} \Delta t\\ 2\Delta t\\ 3\Delta t\\ 4\Delta t\\ 5\Delta t\end{array} \right] \times \dot x(t_1)
+ \left[\begin{array}{c} 1 \\ 1 \\ 1 \\ 1 \\ 1 \end{array} \right] \times x(t_1)
 \]

The optimal control problem is to solve for u[k] matrix in equation (17) subject to the above dynamics, satisfying the boundary conditions defined by the initial and final positions \( (x_0 and x_f) \), respectively, velocities \((\dot x_0  and  \dot x_f \) and accelerations \( \ddot x_0  and  \ddot x_f\).

\begin{equation} -x_{max} \leq x(k\Delta t) \leq x_{max} \end{equation}
\begin{equation} -\dot x_{max} \leq \dot x(k\Delta t) \leq \dot x_{max} \end{equation}
\begin{equation} -\ddot x_{max} \leq \ddot x(k\Delta t) \leq \ddot x_{max} \end{equation}

The quadratic cost function (12), (13) and (14) with the linear equality constraints (17) to (20) define a convex optimization problem.There exist efficient methods for solving problems of this sort, with CVXOPT [11], FORCES [12] and CVXgen [13] presenting techniques for creating C-code based solvers for specific instances of convex optimization problems. Here, solvers are generated using the CVXOPT software of [11], which was able to generate solvers for large problems.
\newline

CVXOPT uses efficient interior point methods tailored to convex optimization problems, as are typical in model predictive control applications, and allows for high-speed implementation with good numerical stability properties.\newline

The generated solvers either return a solution that solves the problem to within some acceptable residuals, or returns that no solution is found. In reality, failure to find a solution can mean that:
\begin{itemize}
\item a solution exists, but the solver failed to find it due to reaching an internal limit; 
\item no solution exists to the conservatively constrained decoupled problem;
\item no solution exists to the fully coupled nonlinearly constrained problem.
\end{itemize}

%

\section{Results}

\subsection{A. Trajectory Tracking \newline}
The proposed algorithm was used to follow particular straight line paths, circles and sinusoidal curves defined by their respective mathematical equations. The results of the above have been shown in Fig (3) (4) and (5). The starting points were taken to be outside the path to follow, but close enough for the circle to find the VTP and start tracking the trajectory. In the plots shown in the figures, the time intervals between every measurement of velocity, acceleration and position was taken to be 0.08s. The radius of the circle used in these plots is 50 units.\newline 

The output of the proposed algorithm was also compared with the trajectory generation model proposed in [7], where the cost function to be minimized calculates the sum of squares of the input jerk. In comparison, the proposed algorithm performs as well as the trajectory generation method proposed in [6].  

The performance of the trajectory tracking is found by solving trajectories from rest, for step times varying from 0.08s to 2s. For each trajectory length, 1000 iterations were executed using python IDLE which took a mean time of 60ms per iteration. 

These were calculated on a PC running Ubuntu 12.04, with an Intel Core i3-330M Dual Core Processor (2.13 GHz) with 4GB RAM. The solver was linked to an executable using Python IDLE. In each case the optimizations were set to maximize speed.

\begin{figure}[h]
  \centering
    \includegraphics[width=0.5\textwidth]{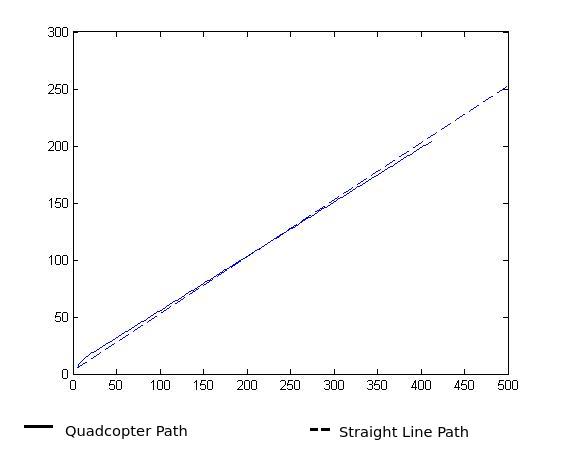}
  \caption{The quadcopter starts at a point outside the linear path. The plot of its predicted trajectory when it tries to trace the line is shown.
}
  \label{fig:Line_follow}
\end{figure}

\begin{figure}[h]
  \centering
    \includegraphics[width=0.5\textwidth]{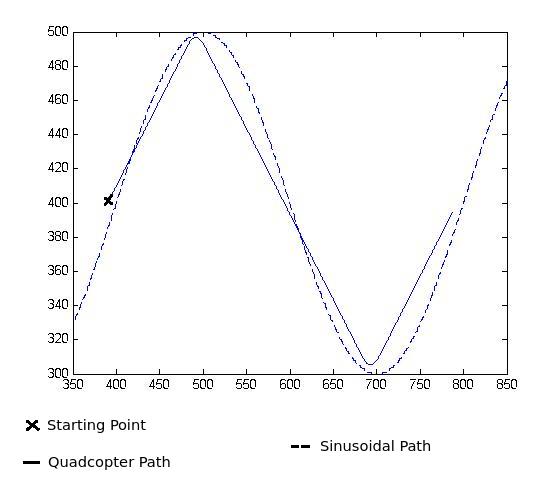}
  \caption{ X intercept vs Y- intercept plot considering that height (Z-axis) is maintained as a constant. In this figure, the predicted trajectory path when it tries to follow a sinusoidal path is plotted.  
}
  \label{fig:Xd_yd}
\end{figure}

\begin{figure}[h]
  \centering
    \includegraphics[width=0.5\textwidth]{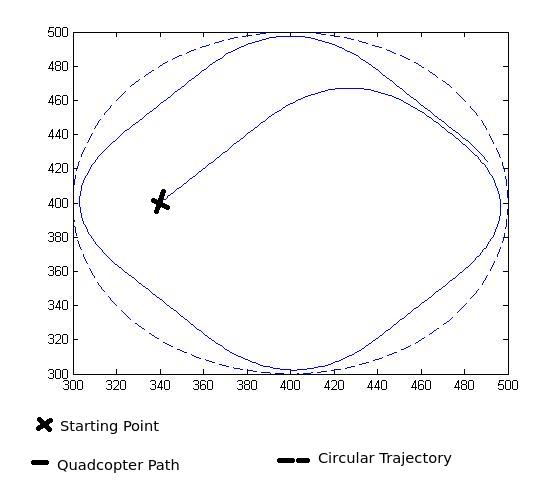}
  \caption{The quadcopter starts at a point inside the circular path. The plot of its predicted trajectory when it tries to follow the circle is shown.
}
  \label{fig:Circle_follow}
\end{figure}

\section{Conclusion}

This paper presents a method for trajectory tracking for quadcopters, from arbitrary initial conditions to a desired end state characterized by a position. The NLGL based destination intercept works satisfactorily in tracking subsequent intercepts in the predefined path.\newline 

The constrained trajectory tracking problem is posed as a convex optimization problem for each decoupled axis, which can be solved in real time. Therefore this technique is suitable for use in feedback as a model predictive controller; it also naturally handles cases where the desired end state evolves over time, such as when trying to hit an uncertain target or when it has to overcome an obstacle, where the target prediction evolves in real time.\newline

The convex optimization problem could be easily modified to achieve different goals, e.g. by removing the equality constraint on end velocity and acceleration, and adding the magnitude of the end velocity to the cost function, one solves a problem somewhat similar to that of [14].\newline

\begin{figure} [h]
\centering
    \includegraphics[width=0.5\textwidth]{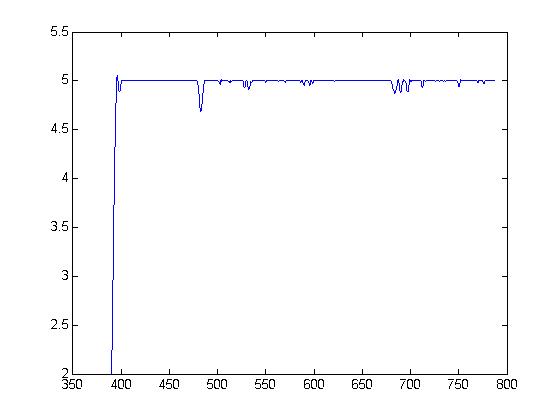}
  \caption{Position Vs Velocity plot of the Decoupled X-axis
}
  \label{fig:xd_vx}
\end{figure}

\begin{figure}[h]
  \centering
    \includegraphics[width=0.5\textwidth]{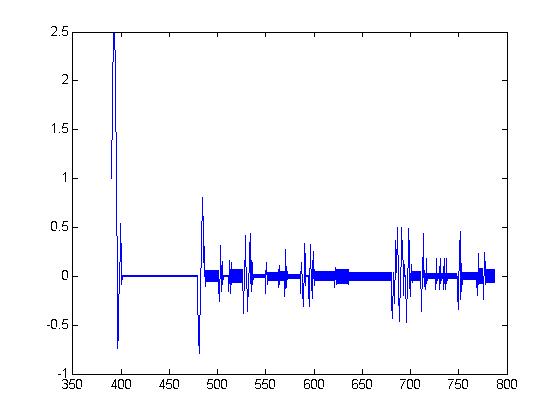}
  \caption{Position Vs Acceleration plot of the Decoupled X-axis
}
  \label{fig:xd_ax}
\end{figure}


\section{Acknowledgments}
This project work was supported by the Robotics Research Centre at International Institute of Information Technology, Hyderabad (IIIT - H).







\section{References}
\begin{enumerate}
\item I. D. Cowling, O. A. Yakimenko, J. F. Whidborne, and A. K. Cooke, “A prototype of an autonomous controller for a quadrotor UAV,” in
Proceedings of the European Control Conference, Kos, Greece, 2007, pp. 1–8.

\item S. Lupashin, A. Sch ̈ llig, M. Sherback, and R. D’Andrea, “A simple learning strategy for high-speed quadrocopter multi-flips,” in IEEE
International Conference on Robotics and Automation, 2010, pp. 1642–1648.

\item S. Weiss, M. Achtelik, M. Chli, and R. Siegwart, “Versatile distributed pose estimation and sensor self-calibration for an autonomous MAV,” in IEEE International Conference on Robotics and Automation, 2012, pp. 31–38.

\item P.B. Sujit, Srikanth Saripalli, Joao Borges Sousa, "Unmanned Aerial Vehicle Path Following - A Survey and Analysis of Algorithms for Fixed Wing UAV", IEEE Control Systems Magazine, Feb 2014, pp. 42 - 59

\item Wang, L. "Model Predictive Control System Design and Implementation using MATLAB", Springer 2009

\item Mark W. Mueller and Raffaello D’Andrea, "A model predictive controller for quadrocopter state interception", 2013 European Control Conference (ECC), July 17-19, 2013, Zürich, Switzerland

\item M. Hehn and R. D’Andrea, “Quadrocopter trajectory generation and control,” in IFAC World Congress, vol. 18, no. 1, 2011, pp. 1485–1491.

\item P. H. Zipfel, "Modeling and Simulation of Aerospace Vehicle Dynamics Second Edition". AIAA, 2007. 

\item Markus Hehn, Raffaello D’Andrea, "Quadrocopter Trajectory Generation and Control"

\item S. Park, J. Deystt, and J. P. How, “Performance and Lyapunov stability of a nonlinear path-following guidance method,” J. Guidance, Control, Dyn., vol. 30, no. 6, pp. 1718–1728, 2007.

\item L. Vandenberghe, "The CVXOPT linear and quadratic cone program solvers"

\item A. Domahidi, A. Zgraggen, M. Zeilinger, M. Morari, and C. Jones, “Efficient interior point methods for multistage horizons arising in receding horizon control,” in Proceedings of the 51st IEEE Conference on Decision and Control, 2012, pp. 668–674.

\item J. Mattingley and S. Boyd, “CVXGen: a code generator for embedded convex optimization,” Optimization and Engineering, vol. 13, no. 1,pp. 1–27, 2012.

\item M. Hehn and R. D. Andrea, “Real-time trajectory generation for in- terception maneuvers with quadrocopters,” in IEEE/RSJ International Conference on Intelligent Robots and Systems, Villamoura, Portugal, 2012, pp. 4979–4984.

\end{enumerate}


\end{article}

\end{document}